# SenSnake: A snake robot with contact force sensing for studying locomotion in complex 3-D terrain

Divya Ramesh, Qiyuan Fu, Chen Li[1]

*Abstract*—**Despite advances in a diversity of environments, snake robots are still far behind snakes in traversing complex 3-D terrain with large obstacles. This is due to a lack of understanding of how to control 3-D body bending to push against terrain features to generate and control propulsion. Biological studies suggested that generalist snakes use contact force sensing to adjust body bending in real time to do so. However, studying this sensory-modulated force control in snakes is challenging, due to a lack of basic knowledge of how their force sensing organs work. Here, we take a robophysics approach to make progress, starting by developing a snake robot capable of 3-D body bending with contact force sensing to enable systematic locomotion experiments and force measurements. Through two development and testing iterations, we created a 12-segment robot with 36 piezo-resistive sheet sensors distributed on all segments with compliant shells with a sampling frequency of 30 Hz. The robot measured contact forces while traversing a large obstacle using vertical bending with high repeatability, achieving the goal of providing a platform for systematic experiments. Finally, we explored model-based calibration considering the viscoelastic behavior of the piezo-resistive sensor, which will for useful for future studies.**

## I. Introduction

Snake robots hold the promise as a versatile platform for traversing in a variety of environments for critical applications [1], [2]. However, despite snakes' remarkable locomotor capacities in 3-D terrain with large obstacles [3]–[5], snake robots still suffer slower speeds and larger slip [6], [7] in similar terrain. Some snake robots use vision to scan the terrain and plan motions to adapt to its geometry [7]–[12]. Others use mechanical [9], [10], [13] or controlled [6], [9], [14] compliance to maintain terrain contact. However, despite progress on using 2-D lateral bending to push against vertical asperities on flat surfaces to generate propulsion [15]–[18], we know little about how to use 3-D body bending to push against complex 3-D terrain to generate propulsion.

An ability to sense and adjust contact forces against the terrain likely contributes to generalist snakes' superior performance. When using 2-D lateral bending to push against vertical structures, generalist snakes can adjust their body bending in real time to maintain contact and control the direction of propulsion [4], [19], suggesting that this is a sensory-modulated process. Snakes possess both cutaneous mechanoreceptors (i.e., skin tactile sensing) and proprioceptors within the muscles and tendons (i.e., internal position, movement, and force sensing) [20]–[24], yet it remains unknown how these sensors are used to detect body position or environmental forces to control locomotion. Such a lack of basic knowledge makes it difficult to study snakes to understand how to sense and control contact forces to generate propulsion in complex 3-D terrain.

Robots have proven very useful as physical models of animals for studying locomotion, especially in complex environments, where it is difficult or impossible to create tractable theoretical models [25]–[28]. In addition, contact force measurements [29] have advanced understanding and performance of many aspects of locomotion and manipulation tasks, including object identification [30], slip detection [31], tactile sensing [32], [33], exploring and interacting with cluttered environments [34], and terrain identification and classification [35], [36]. Here, to establish a robotic platform for studying the physical principles of sensing and controlling contact forces to generate propulsion in complex 3-D terrain and ultimately improve snake robot performance, we developed a snake robot, SenSnake, capable of 3-D body bending with contact force sensing along its body.

Our sensors have two design requirements. First, they must be sufficiently flexible to allow compliant body segments to passively conform to 3-D terrain to improve contact [10]. Second, they must provide sufficient coverage of each segment to accommodate variable contact in complex 3-D terrain and. A variety of sensors have been used to detect forces or contact for snake robots [15]–[18], [37]–[43]. Sensors based on strain gauges [40], [41], optics [42], and switches [18] are rigid. Sensors based on pressure sensitive materials are flexible and more suitable for our needs. Among these, off-the-shelf ones [15]–[17], [38] come in specific sizes and shapes and are less suitable for high coverage; custom ones [37], [39], [43] provide high coverage but require substantial effort and special equipment to manufacture. Here, we chose a low-cost piezo-resistive pressure sensor design that can be custom made to any shape for maximal coverage and is easy to manufacture, following recent work [30], [44].

We first developed and tested an initial robot prototype (Sec. II). Based on limitations revealed from the testing, we refined the robot and sensor design (Sec. III). This enabled the robot to move over a large obstacle and measure contact forces with high repeatability (Sec. III), achieving our major goal of providing a robotic platform for systematic experiments (rather than optimizing robot performance as in many robotics studies). In addition, we performed experiments to calibrate the sensors on the refined robot, which can improve force estimate accuracy and inform future design of terrain testbeds for studying complex 3-D terrain traversal (Sec. IV). Finally, we summarize contributions and discuss future work (Sec. V).

[1]Dept. of Mechanical Engineering, Johns Hopkins University, Baltimore, MD 21218 USA [redacted]. Corresponding author: Chen Li. This work was supported by an Arnold and Mabel Beckman Foundation Beckman Yong Investigators Award, a Burroughs Wellcome Fund Career Award at the Scientific Interface, a Johns Hopkins University Catalyst Award, and the JHU Whiting School of Engineering start-up funds.

## II. INITIAL SENSOR & ROBOT DEVELOPMENT

### A. *Initial robot prototype*

The initial robot prototype, SenSnake v1 (0.9 m long, 0.044 m cross-sectional radius, 4.6 kg), had 12 alternating pitch and yaw segments to bend in 3-D (Fig. 1A). Each segment had a servo motor (Dynamixel XM430-W350-R) fully enclosed in a soft shell casted from silicone (Ecoflex 00-30, Smooth-On Inc.) (Fig. 1C, black) attached via 3-D printed shell holders (Fig. 1C). On the outside of the soft shell, each pitch segment had an array of four sensors (in a 2 × 2 arrangement) on its bottom, whereas each yaw segment had two such arrays on both sides (Fig. 1D). This resulted in a total of 72 sensors, which could be recorded at a sampling frequency of 17 Hz (see Sec. II, C). The soft shell was intended to deform passively during terrain interaction to improve terrain-sensor contact.

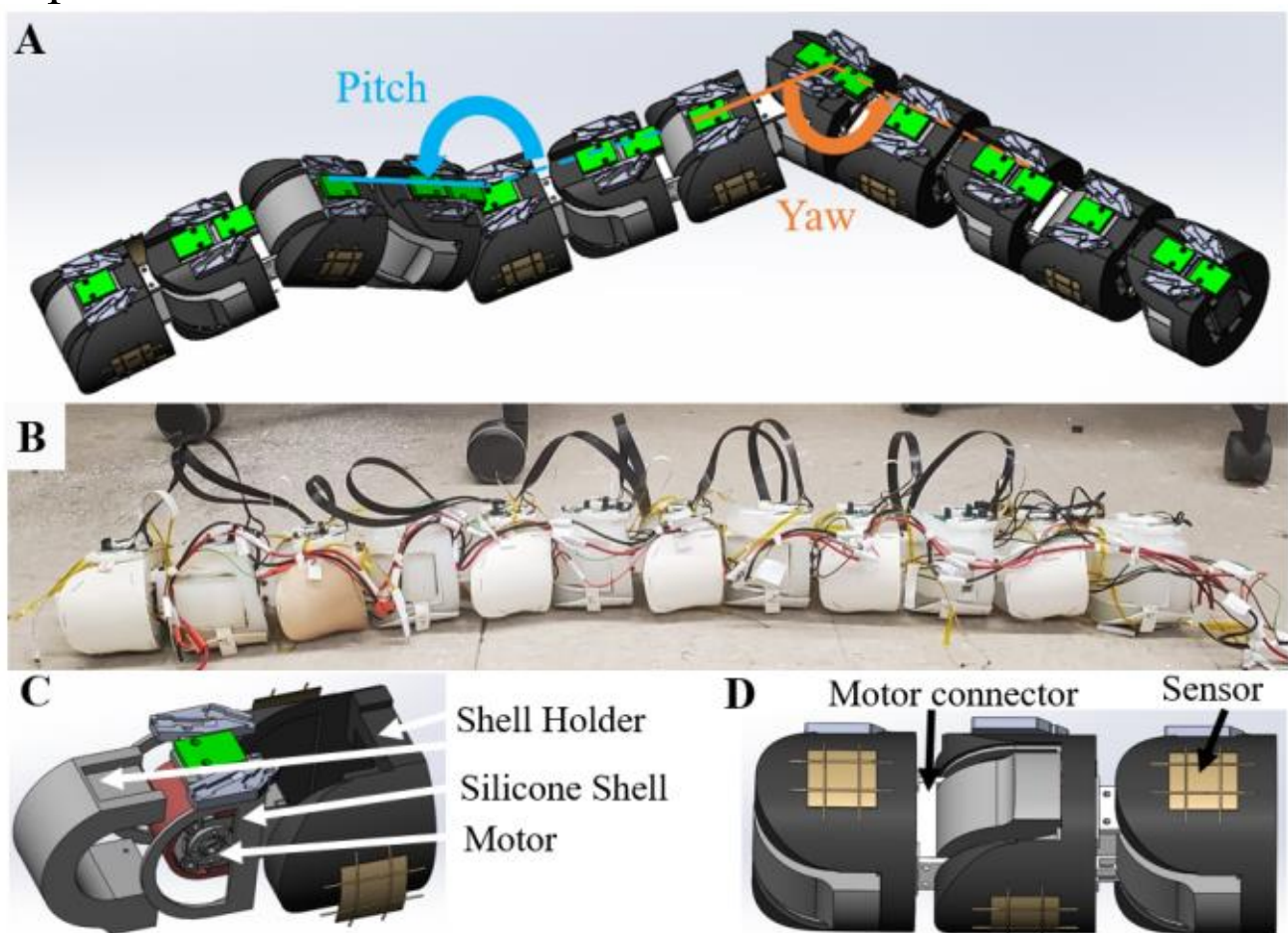

**Fig. 1. SenSnake v1.** (A) CAD showing 3-D bending with pitch and yaw degrees of freedom. (B) Photo of robot without sleeve. (C) Exploded view of a segment. (D) Side view CAD of three segments with sensors on the side of yaw and bottom of pitch segments.

The robot was powered by a 12 V DC power supply. The motors were daisy-chained and controlled via a USB communication convertor (U2D2, Dynamixel). A rubber layer covered each sensor to prevent sensor wear and tear (Fig. 1B). A PolyEthylene Terepthalate braided sleeve (Flexo Pet, Techflex) covered the robot to reduce friction. Eighteen 8-pin FFC cable adapter to 8-DIP adapter PCB boards (green in Fig. 1A, C) were attached to the top part of the shell holder to connect sensors to a Data Acquisition board (DAQ).

### B. *Sensor fabrication*

The sensor array consisted of seven layers (Fig. 2A). A piezo-resistive film (3M Velostat, 0.1 mm thickness, Adafruit Industries) was sandwiched between parallel stainless steel conductive threads (3 ply, 0.25 mm thickness, Adafruit Industries) above and below, placed perpendicular to each other (Fig. 2A). A crossing of thread above and below forms a single sensor. An adhesive sheet (Gizmo Dorks 468MP, 2.54 mm thickness, 3M) was placed over the conductive threads (Fig. 2A) and adhered to the piezo-resistive film, followed by a sheet of plastic wrap.

Each sensor experiences a reduction in resistance $R_s$ on application of normal force. The sensor conductance, $C_s = 1/R_s$, increases linearly with the force applied (Fig. 2B):

$$C_s = mF + d \tag{1}$$

where $m$ and $d$ are constants (multimedia material, video 1).

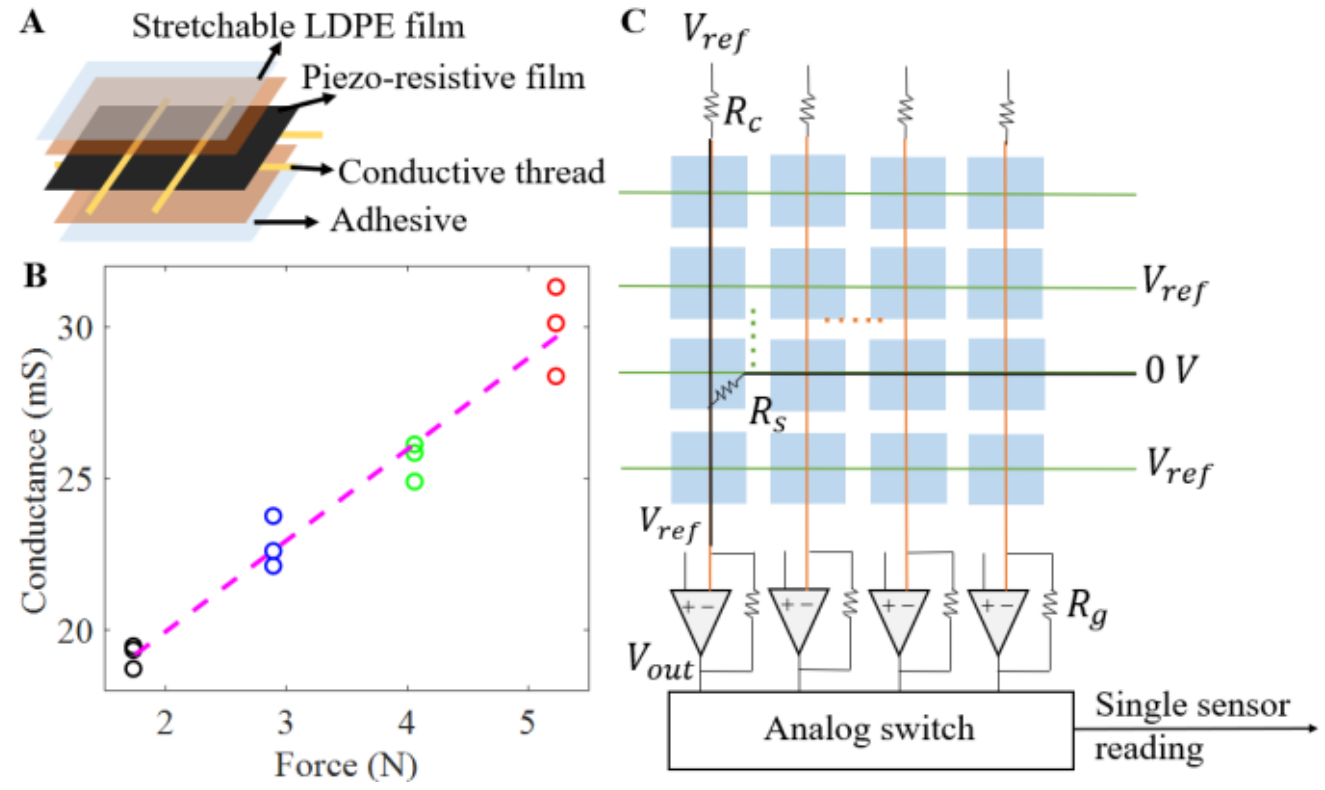

**Fig. 2. Sensor fabrication, calibration, and data acquisition.** (A) Piezo-resistive sensor design. (B) Average conductance between 225 s and 300 s as a function of force applied (3 trials each) from calibration experiments in Sec. IV. Dashed line is best linear fit. (C) Signal isolation circuit to collect the sensor readings. Blue squares are piezo-resistive sheets. $V_{ref}$ = 2.5 V, $R_g$ = 1000 Ω, $R_C$ =900 Ω.

### C. *Sensor data acquisition*

To obtain sensor resistance $R_s$, we replicated the DAQ designed in [30], which uses a signal isolation circuit to scan through sensors one at a time using a multiplexer and a demultiplexer (Fig. 2C) and minimize sensor crosstalk, an undesirable effect that sensors close to each other affect each other's resistance [45]. We characterized the sampling frequency of the DAQ, defined as the frequency at which data from all sensors being tested can be received. As expected, sampling frequency decreased monotonically with the number of sensors (Fig. 3A), starting at about 20 Hz for one sensor, decreasing to 10 Hz for ~300 sensors, and approaching only a few Hz for ~1000 sensors. When two DAQ were used, sampling frequency decreased but only slightly (Fig. 3B).

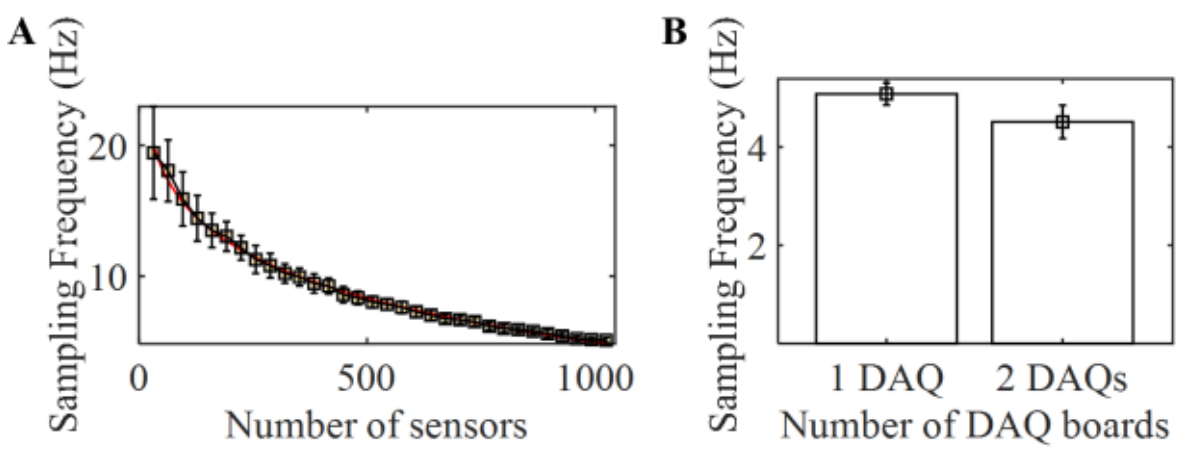

**Fig. 3. Sampling frequency of sensor data acquisition.** (A) Sampling frequency as a function of the number of sensors (mean ± s.d. over 100 sampling cycles). (B) Comparison of sampling frequency between one and two DAQ used. 1024 sensors were scanned for each case.

### D. *Experiments and issues revealed*

We tested SenSnake v1 on flat rigid ground and a pile of small wooden blocks using lateral undulation (15.24 cm long, 5.08 cm wide, 3.38 cm tall: Fig. 4A-B). In both cases, the robot did not progress forward due to a lack of anisotropic friction necessary for undulating on smooth flat surfaces [46]. We observed constant forces when the robot remained stationary for the first 10 s (Fig. 4C-D). During lateral undulation, forces oscillated periodically on flat ground (Fig. 4C) and not so regularly on blocks (Fig. 4D). These results showed that the sensors can detect the expected forces during locomotion (multimedia material, video 2).

However, these tests also revealed several issues. First, we observed small signals on a large portion of force sensors on the body segments in contact with the block pile. This suggested that the silicone shell did not deform sufficiently to distribute highly localized stresses at terrain contact points widely to reach the exact locations of the sensors. Because each sensor covered only a small area at the crossing of the perpendicular conductive threads, it could not detect a large force signal. Contrary to expectation, the differences between the four sensors within each 2 × 2 sensor array on the block pile were similar to those on flat ground (Fig. 4C-D) likely a result of the lack of direct contact at the sensor point. This further showed that the intended high sensor spatial resolution did not outweigh the small sensor area limitation. Moreover, some of the sensors developed substantially noisy reading during experiments (Sensors 3R, 4B, 11L in Fig. 4C-D) because the sensor-chipboard connection wires became loose.

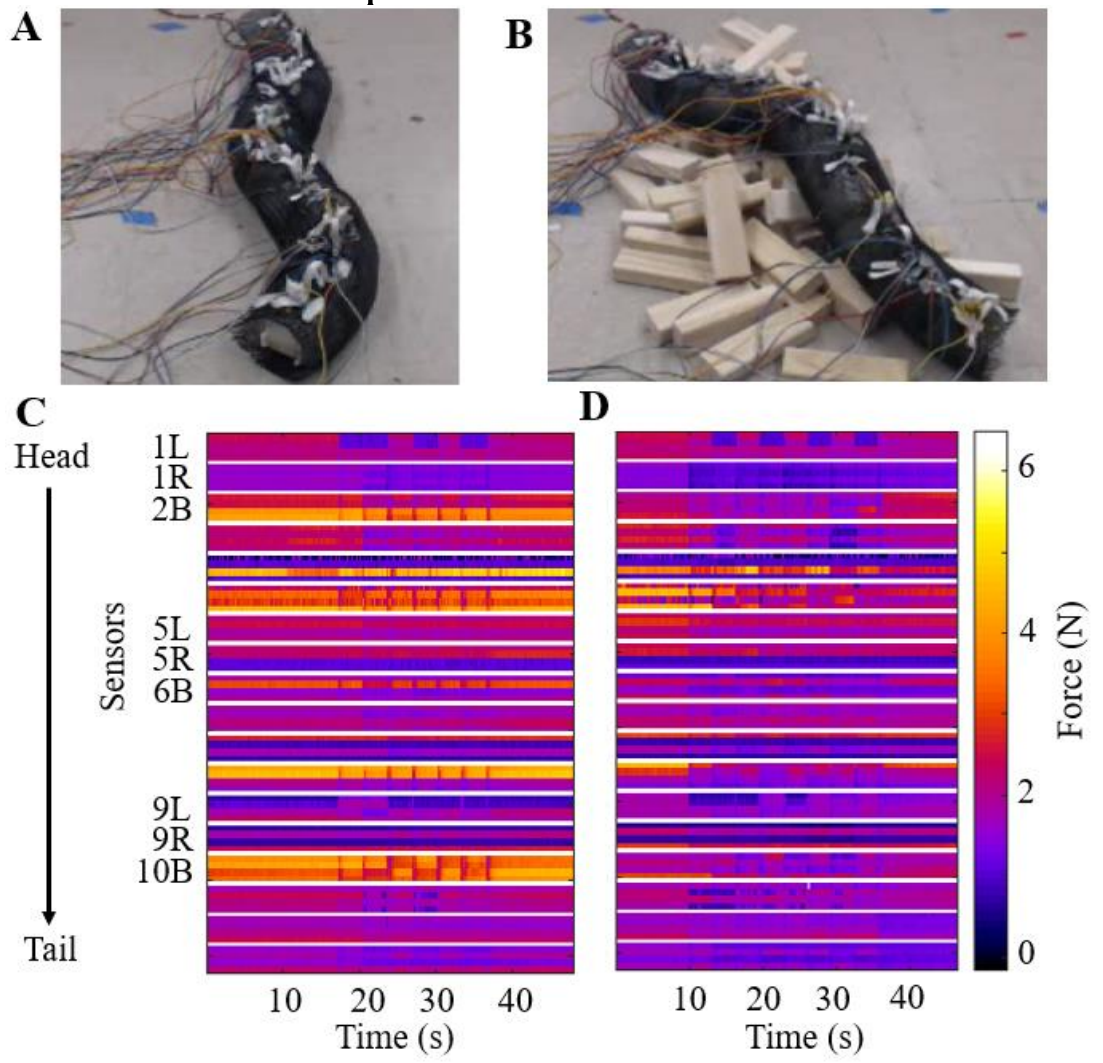

**Fig. 4. Initial robot prototype experiment.** (A) Lateral undulation on flat ground. (B) Lateral undulation on rubble of wooden blocks. (C) Force readings varying with time on flat ground. (D) Force readings varying with time on rubble of wooden blocks. 1L and 1R indicate 1st segment left and right, 2B indicates 2nd segment bottom.

We also tested the robot traversing a single large obstacle as high as 0.28× robot length by propagating a vertical bending shape that conforms to the obstacle down its body, a strategy inspired by recent animal observations [47]. Although the robot was able to generate the desired shape evolution on its own, it failed to use it against the large obstacle to propel forward. Examination of motor angle data revealed that the motors could not reach the desired positions during obstacle interaction. This was likely because pushing against the large obstacle resulted in high contact forces concentrated on segments contacting the forward half of the obstacle. To propel the entire robot forward, these segments must sustain these large contact forces to overcome the large frictional drag from the substantial robot weight from the silicone layers. This large force requirement, together with restrictions from the sleeves, probably resulted in motor overload and trigged motor to give to prevent damage.

## III. Refined Sensor & Robot Development

We made several design and fabrication improvements to address these issues in a refined robot. These include: (1) replacing the solid silicone shell with a more compliant, hollow shell to reduce robot weight and sleeve restriction and improve body/sensor-terrain conformation; (2) replacing each 2 × 2 sensor array with a single sheet sensor [44] for more reliable force detection, further increasing sampling frequency; (3) adding a sensor to the left, right, and bottom sides of each segment to improve overall body sensor coverage; and (4) embedding wiring inside the robot to minimize disturbance during locomotion.

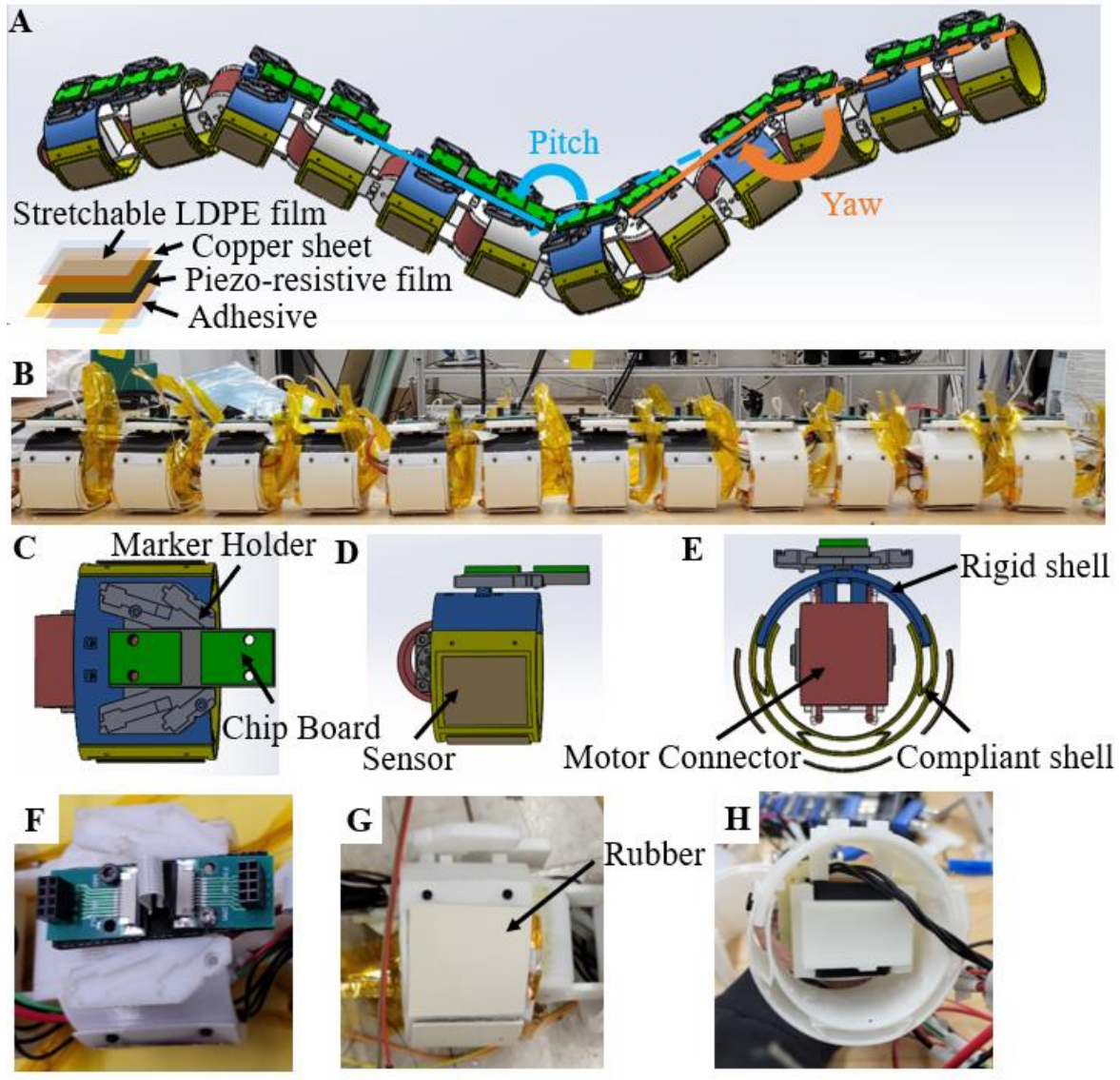

**Fig. 5. SenSnake v2 design.** (A) CAD showing 3-D bending of robot. (B) Robot photo. Top, side, and front view CAD (C-E) and photos (F-H) of a segment.

### *A. Refined sensors*

We switched to a piezo-resistive sheet sensor to reduce the number of sensors, improve wire packaging, and increase sensor area. The sheet sensor is similar to the sensor array in design and working principle, except that the conductive threads (Fig. 2A) were replaced with a copper conductive sheet (Copper foil sheet with conductive adhesive, 0.07 mm thickness, Adafruit Industries) on either side (Fig. 5A).

### *B. Refined robot*

The refined robot, SenSnake v2 (0.96 m long, 0.036 m segment radius, 2.4 kg), has 12 segments (Fig. 5B) with same joint structures as the initial robot (Fig. 5A). To make the robot lighter, the silicone shell was replaced with a compliant, hollow shell to evenly distribute the force on the sensor (Fig. 5E). Because only the sides and bottom of the robot came in contact with the terrain obstacles during traversal, the rigid upper part of the shell was 3-D printed using PLA (Fig. 5C, blue), whereas the rest of the shell was 3-D printed soft using TPU (Fig. 5D, yellow).

Three sheet sensors (4 cm long, 3.5 cm wide) were distributed over majority of the soft shell (Fig. 5E) to detect forces on the left, right and bottom sides of each segment, totaling 36 sensors with a sampling frequency of 30 Hz. A rubber layer covered each sensor to prevent sensor wear and tear (Fig. 5G). We improved wiring to be enclosed inside the shell for better protection (Fig. 5F, H). We installed 3-D printed holders to mount LED motion capture markers to track each segment.

## C. *Experiments*

To test how well our sensor and robot improvements solved the problems in the initial prototype and demonstrate its usefulness for understanding locomotion in complex terrain, we tested the refined robot on wooden half-cylindrical obstacle constructed from assembling laser cut boards (Fig. 6A-B). To reduce friction, we covered the entire surface with plastic sheet (0.254 mm polytetrafluroethylene sheet, McMaster, USA). Eight motion capture cameras (PhaseSpace IMPULSE X2) tracked 4 unique LED markers on each segment to obtain 3-D kinematics at 960 Hz (Fig. 5C).

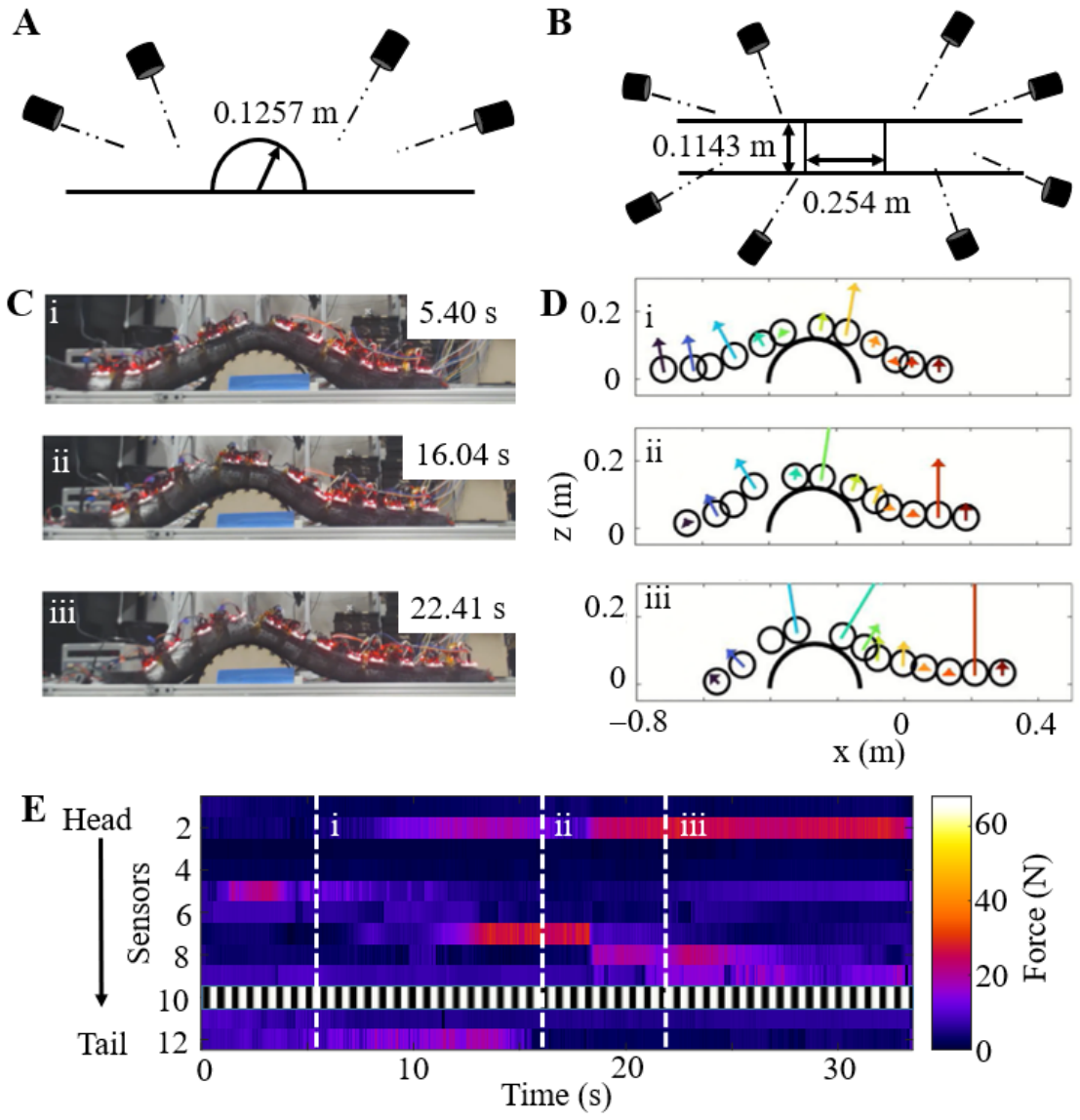

**Fig. 6. Force response generation during robot vertical bending.** (A, B) Experimental setup. (C) Side view snapshots of robot traversing a large half-cylindrical obstacle. (D) Side view reconstruction of segment positions from motion capture data. (E) Measured force as a function of time for all sensors on bottom of the robot. White lines in B correspond to snapshot times in A. We note that sensor 10 does not give reliable force measurement due to loose sensor connection. Each sensor is individually calibrated (Eqn. 1). Each sensor's force data were offset by the absolute value of minimal negative value to remove artifacts of "negative pressure" from small sensor drift due to disturbance from the robot's self-deformation.

We used feedforward control using Robot Operating System at a frequency of 50 Hz to propagate a pre-defined vertical bending shape down the body at 0.034 rad/s in a follow-the-leader manner. This pre-defined shape was generated by manually pushing the robot down to conform to the obstacle and recording the motor angles. The controller used linear interpolation of motor angles over time to propagate the shape down the body. The robot started with the 6$^{th}$ segment on top of middle of the obstacle, because on a flat ground a vertical bending could not generate sufficient propulsion to move it forward.

Overall, the robot conformed well to the obstacle and generated sufficient propulsion to propel itself forward to traverse the large obstacle. As the robot moved forward, all the contact forces patterns propagated backward relative to the robot (Fig. 6E). For the first 10 seconds after the robot started moving, forward motion was smooth (Fig. 6Ci), with substantial normal forces (~5 N, ~22% robot weight) on the segments contacting the front of the obstacle and the horizontal surface (Fig. 6Di). Besides supporting part of the robot weight, the normal force against the front side of the obstacle also resulted in forward propulsion.

Until the middle of the robot passed over the middle of the obstacle (Fig. 6Ci), the robot slowed down momentarily on the obstacle, presumably due to a relative contact. As the robot continued to propagate bending backward, the segment contacting the front side of the obstacle pushed harder and generated a very large force (23 N, 98% robot weight) (Fig. 6ii). This buildup of forward propulsion eventually helped the robot overcome frictional drag and slip forward rapidly, after which it resumed steady motion (Fig. 6C-Eiii). See multimedia material, video 3 for an example video.

We performed three trials and found excellent repeatability in the robot's motion and sensor data (Fig. 7A-B), achieving our main goal of providing a robotic platform for systematic experiments to understand principles [25]–[28].

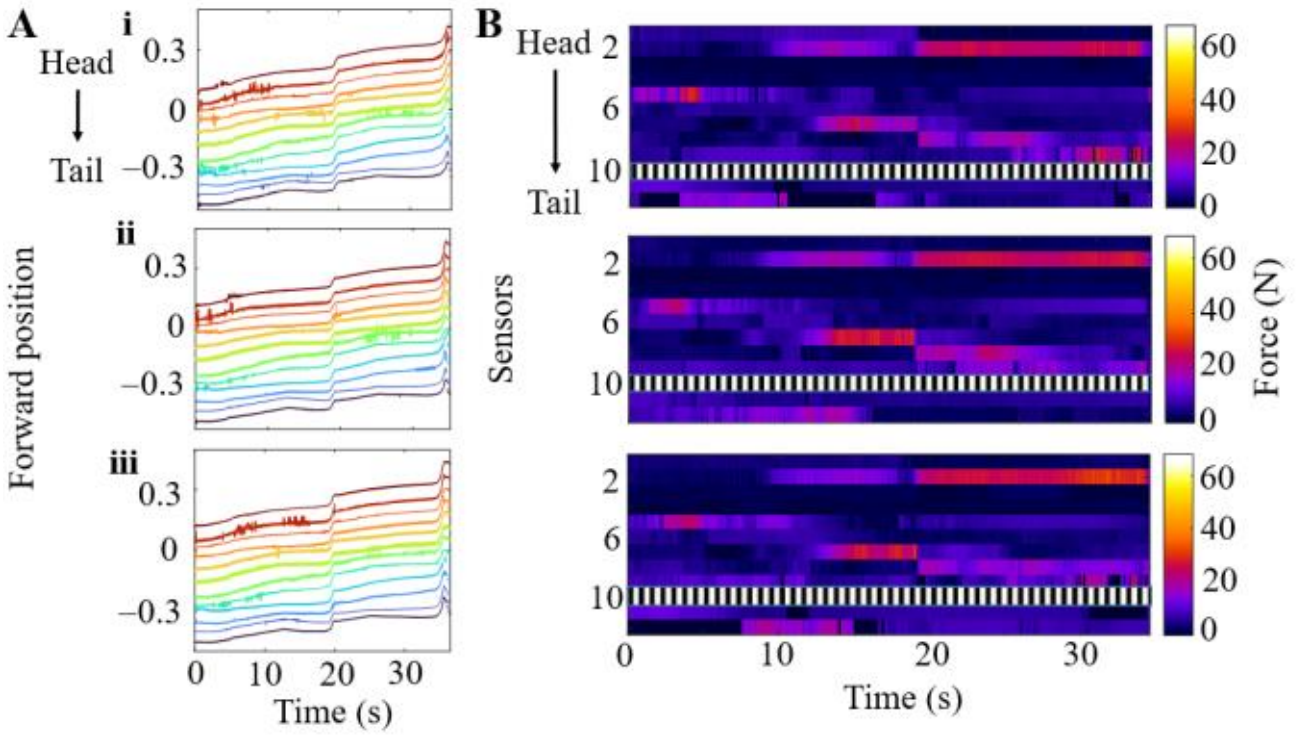

**Fig. 7. High system repeatability for robophysics studies.** (A) Segment forward position as a function of time. (B) Force as a function of time for all bottom sensors. Sensor 10 does not give reliable force measurement due to loose connection.

# IV. Model-Based Sensor Calibration

With proper design and fabrication, piezo-resistive force sensors enabled our snake robot to detect contact forces when traversing large obstacles. This is consistent with previous success in snake robots doing so against large vertical structures on horizontal surfaces [15]–[18]. One distinction is that these previous studies focused mainly on using sensor data to generate locomotion in a robot. Our robot sensor development is not only to help generate locomotion, but also to provide quantitatively accurate measurements necessary for understanding the physical principles of snake propulsion generation using 3-D body bending.

For our goal, it is useful to consider the viscoelastic nature of piezo-resistive material, which causes a creep behavior of the sensor reading after force application [48]. We carried out sensor modeling and calibration experiments to obtain high-fidelity force information from sensing reading, following recent work in developing a physics-based sensor model [44].

## A. *Sensor model*

Various models have been developed to model creep behaviors in viscoelastic materials under a constant stress [48]. Simpler 2-parameter solid models such as Maxwell or Kelvin-Voigt models do not accurately describe creep or relaxation, respectively [48]. Here, we used the 3-parameter Kelvin-Voigt representation of standard linear solid model (Fig. 8A) [44], [48], which was demonstrated to accurately describe creep and relaxation in piezo-resistive material [44]. The stress-strain dynamics from this model is:

$$\sigma + \frac{\mu_1}{E_0+E_1}\dot{\sigma} = \frac{E_0 E_1}{E_0+E_1}\varepsilon + \frac{\mu_1 E_0}{E_0+E_1}\dot{\varepsilon} \tag{2}$$

where $\sigma = F/A$ is the stress applied, $\varepsilon$ is the induced strain, $E_0$ and $E_1$ are elastic coefficients and $\mu_1$ is viscous coefficient of the piezo-resistive material, $F$ is the force applied, and $A$ is the sensor's active area that is being deformed. Note that $E_0$, $E_1$, and $\mu_1$ are model fitting parameters that linearly increase with the force applied [44], not constant material properties.

In addition, a previously developed physics model well describes the physical mechanism of how sensor deformation leads to resistance change [44]. The resistance-strain relationship from this model is:

$$R_s = \frac{\rho_1+\rho_2}{2}\sqrt{\frac{\pi H}{F}} + R_0(1-\varepsilon)e^{-\gamma D\varepsilon\left[\left(\frac{\pi}{6\phi}\right)^{\frac{1}{3}}-1\right]} \tag{3}$$

where $R_s$ is the total resistance measured, $\rho_1$ and $\rho_2$ are the resistivities of the piezo-resistive and conductive materials, respectively, $H$ is the hardness of the material that measures the material's resistance to localized plastic deformation, $R_0$ is the initial resistance of the piezo-resistive material (2.58 kΩ), $D$ is the filler particle diameter (500 nm), and $\phi$ is the volume fraction of filler particles (0.2873). $\gamma$ is defined by:

$$\gamma = \frac{4\pi}{h}\sqrt{2m_e\varphi} \tag{4}$$

where $h$ is the Plank's constant, $m_e$ is the mass of an electron, and $\varphi$ is the potential barrier height between two adjacent filler particles (0.05 eV). Parameter values are from [44].

### B. Model parameter estimation

The sensor model parameters $E_0$, $E_1$ and $\mu_1$ are estimated using the least squares parameter estimation method [49]. The strain $\varepsilon$ is estimated using the Eqn. 3 for a given constant force and the measured $R_s$. The stress-strain dynamics in Eqn. 2 can be rewritten as:

$$\sigma + a\dot{\sigma} = b\varepsilon + c\dot{\varepsilon} \tag{5}$$

where:

$$a = \frac{\mu_1}{E_0+E_1},\ \ b = \frac{E_0 E_1}{E_0+E_1},\ \ c = \frac{\mu_1 E_0}{E_0+E_1} \tag{6}$$

The stress-strain system can be further rearranged:

$$\sigma = \phi\Theta, \phi = [\varepsilon \quad \dot{\varepsilon} \quad -\dot{\sigma}] \tag{7}$$

If $\phi$ is a non-singular matrix, then the following equation can be used to estimate Θ and the model fitting parameters:

$$\widehat{\Theta} = (\phi^T\phi)^{-1}\phi^T\sigma = [\hat{b}\ \hat{c}\ \hat{a}]^T \tag{8}$$

An exponential fit of the sensor conductance $C_{fit}$ was used to estimate the sensor model parameters:

$$C_{fit} = c_1 + c_2 e^{-c_3 t} \tag{9}$$

We estimated $c_1$, $c_2$ and $c_3$ for each trial by finding the least root mean square error fit while constraining their ranges so that the fit visually matched maximal and minimal conductance values.

### C. Calibration setup

We developed a calibration system to calibrate the sensors systematically and repeatedly. A servo motor rotates a 3-D printed wheel (Fig. 8D, orange), which tries to rotate another wheel through a cable with a spring (stiffness = 246 N/m) to push a probe against the sensor. The spring allows the pushing wheel (Fig. 8D, red) to stop rotating while generating a controlled force that can be measured by measuring spring deformation. During calibration, we attached a fully assembled segment onto an aluminum beam and actuated the motor to apply a constant force for 370 s at a sampling frequency of 6 Hz. We tested four different constant forces, 1.75, 3, 4 and 5.25 N, and collected 3 trials each (multimedia material, video 4).

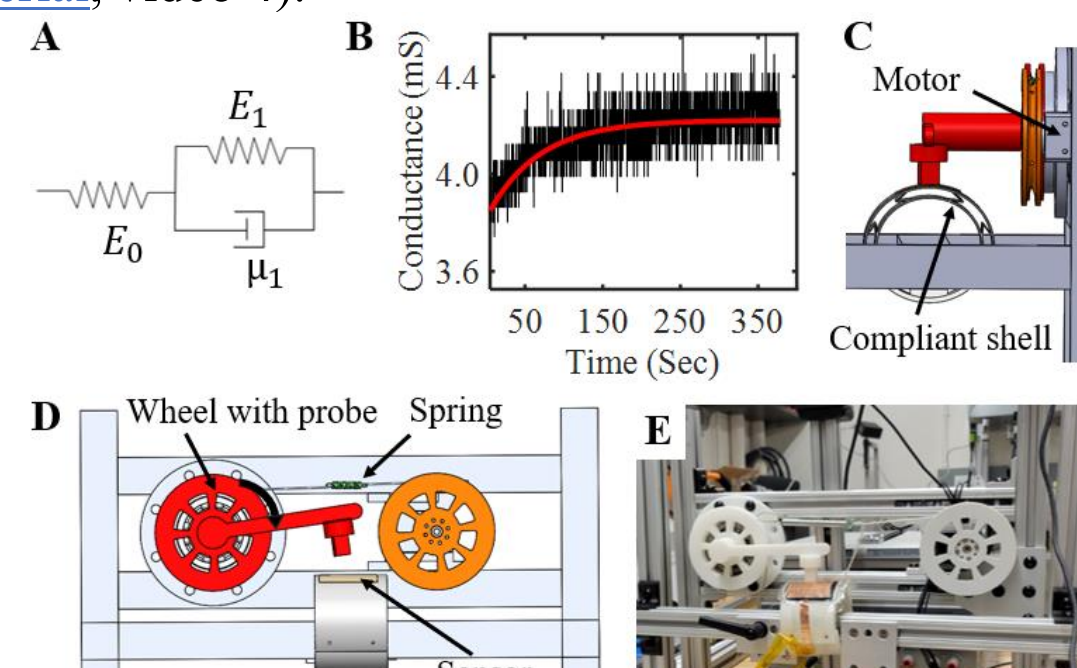

**Fig. 8. Calibration setup.** (A) Sensor model. (B) Sensor conductance vs. time (black) and exponential fit (red) using Eqn. 9. (C, D) Front and side view schematics of calibration setup. (E) Front view photo of calibration setup.

### D. Choice of calibration probe

In complex 3-D terrain, the robot may push against various objects, resulting in flat surface (Fig. 9A-B), edge (e.g., Fig. 9C), corner, or point (Fig. 9D) contact. These diverse contact conditions may affect the repeatability of the sensors and fidelity of sensor model, but few studies considered their effects [44], [50]. To test how robust our sensors and model-based calibration is, we tested four probes simulating different types of contact: a large flat probe covering the entire sensor (Fig. 9A), a small flat probe (Fig. 9B), a sharp edge probe (Fig. 9C), and a multi-point probe (Fig. 9D).

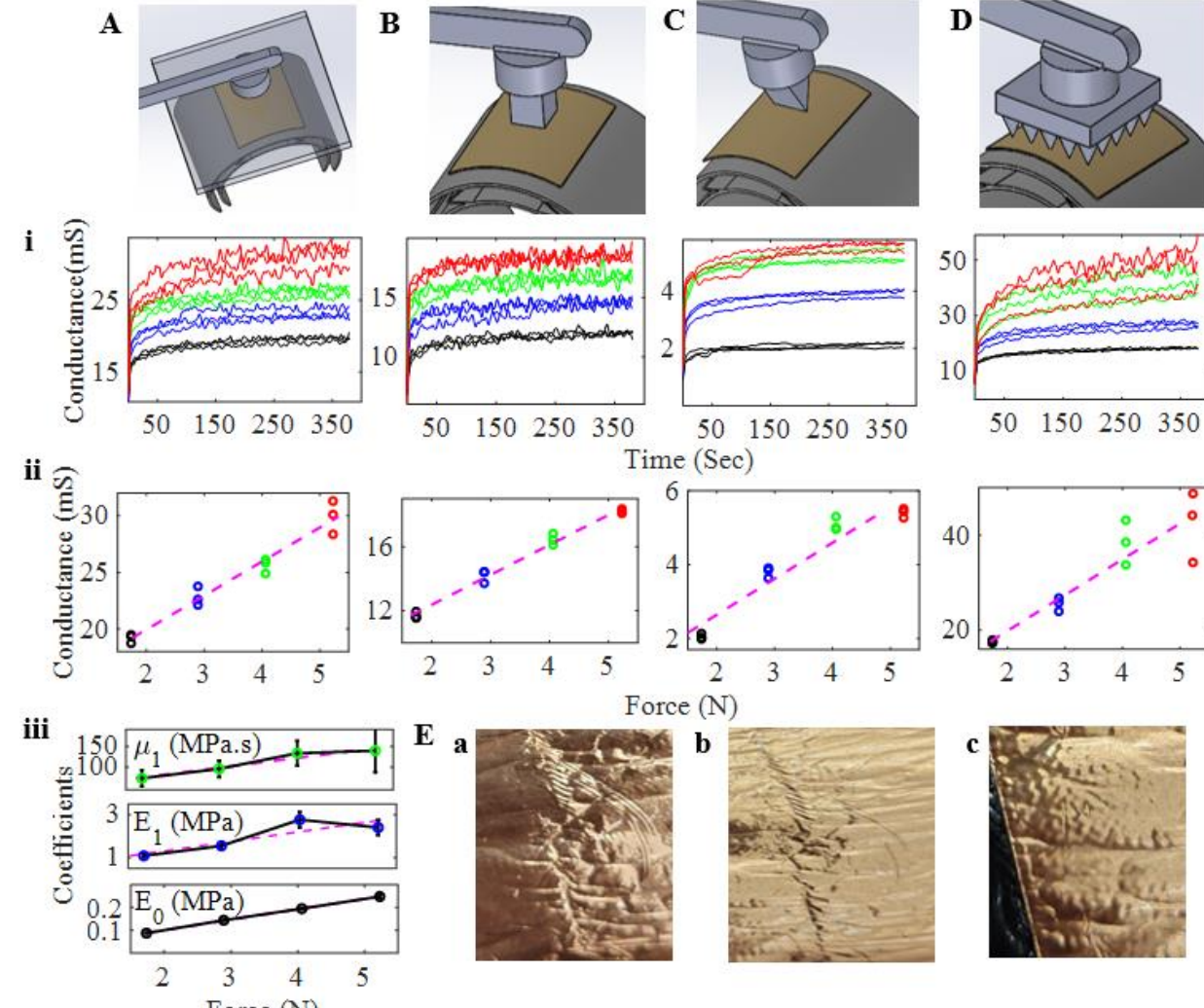

**Fig. 9. Model-based sensor calibration results.** (A) Large flat probe. (B) Small flat probe. (C) Sharp edge probe. (D) Multi-point probe. (i) Measured sensor conductance as a function of time for four different constant forces. (ii) Average conductance after 300 s using data in (i) as a function of applied force. (iii) Model fitted sensor parameters (Eqn. 6) as a function of applied force using large flat probe. (E) Damaged sensor from multiple trials using probes in B-D. The dashed lines in (ii) and (iii) indicate linear fit lines. Black, blue, green, and red are for applied forces of 1.75, 3, 4, and 5.25 N.

For the large and small flat probes, all three estimated parameters increased linearly with force (Fig. 9A, B, iii), consistent with previous observations [44]. However, for the sharp edge and multi-point probes, the estimated parameters increased less linearly as force increased (Fig. 9C, D, ii). In addition, the piezo-resistive sensor layer creased substantially after repeated calibration tests with the small flat, sharp edge, and multi-point probes (Fig. 9E). Close observations of probe-sensor interaction during calibration showed that the local shape of the compliant shell changed significantly with forces concentrated on small contact areas, likely contributing to sensor creases (multimedia material, video 5).

These observations are informative for our future systematic robophysics experiments. To ensure high-fidelity force data to gain principled understanding of locomotion in complex 3-D terrain, it is more practical to design terrain testbeds with large obstacles that are sufficiently smooth to minimize edge or corner contact so that the resulting forces can be well described by the sensor model. Certainly, sensors more robust to such contacts inevitable in the real world still need to be developed for robotic applications.

### *E. Dynamic force measurement using the sensor model*

The piezo-resistive sensor's creep occurs with a characteristic time of $10^2$ s (Fig. 8B), longer than typical periods ($10^{-1}$-$10^1$ s) of most robot locomotion (although soft robots can be as slow as this creep behavior [51]). Because of this, creep behavior was not considered in most previous mobile robot studies with piezo-resistive force sensors [17], [35]–[37], [52]. In preliminary experiments, we found that the snake robot can easily become stuck when attempting to move in complex 3-D terrain resulting in sustained contact. In this case, considering sensor creep behavior is necessary for estimating dynamic forces accurately.

To do so, we can decompose dynamic forces into multiple infinitesimal phases and fitting the sensor model to each phase, using the Boltzmann superposition principle [48], [53] (Fig. 10A):

$$\varepsilon(t) = \sigma_0 J(t) + \int_{0^+}^{t} J(t-\tau)\frac{d\sigma(\tau)}{d\tau}d\tau \qquad (10)$$

where $\sigma_0$ is the initial stress applied at $t = 0$, $\tau$ is the time that measures dynamic changes in stress $\sigma(t)$ (which does not exist for a constant force), $0^+$ denotes the time after the initial stress is applied, and $J(t)$ is defined as:

$$J(t) = \frac{\varepsilon(t)}{\sigma(t)} = \frac{1}{E_0} + \frac{1}{E_1}(1 - e^{-\frac{E_1}{\mu_1}t}) \qquad (11)$$

For example, the resulting strain for the force in Fig. 10A is (where $A$ is sensor active area):

$$\varepsilon(t) = \frac{F_0}{A}J(t) + \frac{F_1 - F_0}{A}J(t-t_1) + \frac{F_2 - F_1}{A}J(t-t_2) + \cdots \qquad (12)$$

We first tested how well this works for a simple 2-step force input (Fig. 10B, black) measured from the calibration setup. We applied the model (Eqns. 2 and 3) using the superposition principle (Eqn. 10) to estimate the resulting sensor conductance (Fig. 10 C, red), which well matched the measured conductance (Fig. 10 C, black). We then applied the model using the superposition principle in the reverse direction, using the measured sensor conductance (Fig. 10C, black) as input to estimate the force applied (Fig. 10B, red), which well matched the measured force (Fig. 10B, black).

Next, we tested how well this works for a dynamic, sinusoidal force input (Fig. 10D, black), which is generated mathematically and free of noise. We applied the model using the superposition principle to estimate the resulting conductance (Fig. 10E, red), which is also noise-free. Next, we added Gaussian noise to the resulting conductance (Fig. 10E, black) to simulate real measured conductance data. Then we applied the model using the superposition principle in the reverse direction, using the simulated sensor conductance (Fig. 10C, black) as input to estimate the force applied (Fig. 10D, red). Despite the noise, it well matched the sinusoidal force input (Fig. 10D, black).

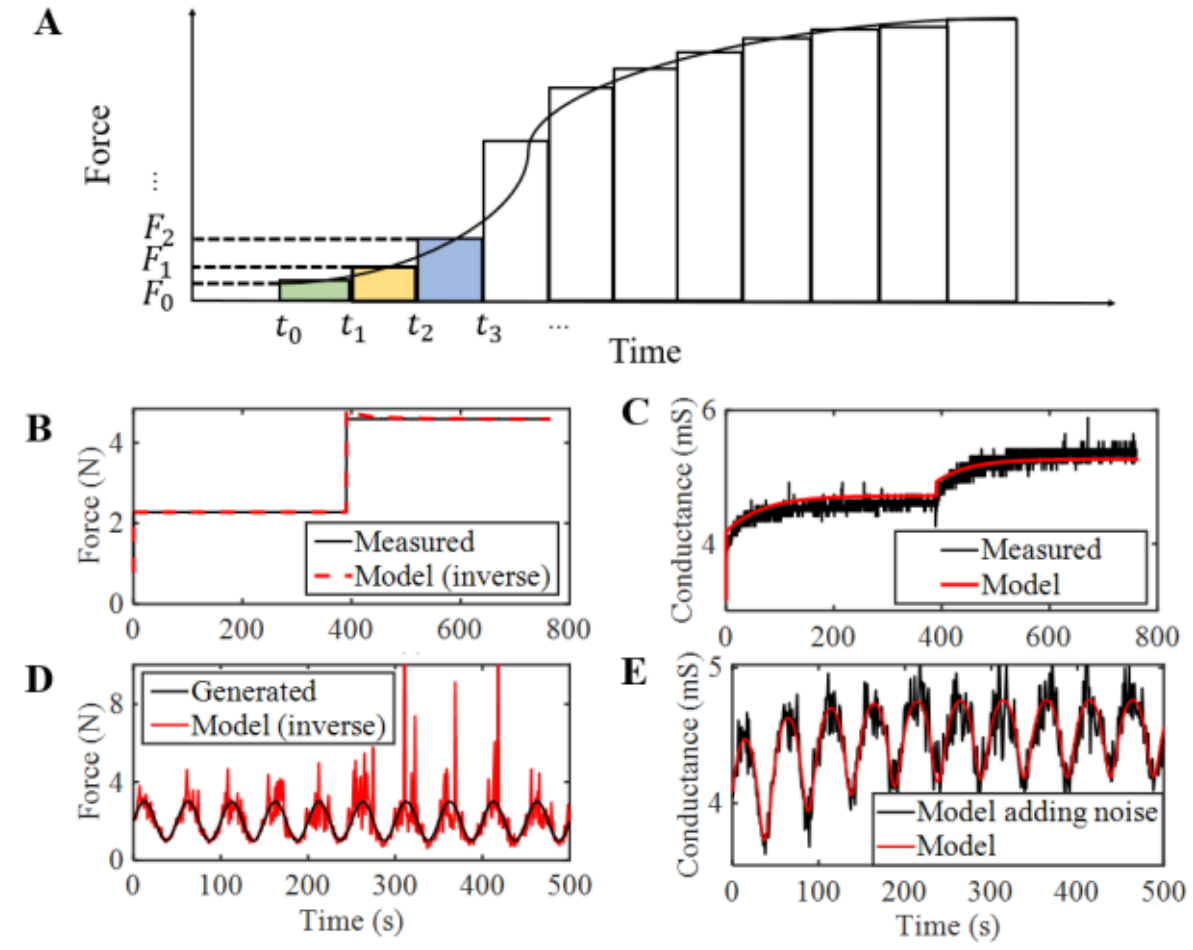

**Fig. 10. Proof of concept of model-based dynamic force estimation.** (A) Idea of dynamic force decomposition using Boltzmann superposition principle. (B, C) Proof of concept using 2-step force input. (B) 2-step force measured using calibration setup (black) vs. estimated from model (red). (C) Sensor conductance measured (black) vs. estimated (red) using 2-step force. (D, E) Proof of concept using sinusoidal force. (D) Sinusoidal force input generated mathematically (black) vs. estimated from model (red). (E) Sensor conductance estimated using sinusoidal force (red), with Gaussian noise added (E, black) to simulate real data.

## V. SUMMARY & FUTURE WORK

To provide a platform for studying how to use contact force sensing to modulate 3-D body bending to propel against 3-D terrain for locomotion, we developed a snake robot with contact force sensors distributed along its entire body. Through two development and testing iterations, our robot was able to obtain contact force measurements while moving over a large obstacle with high repeatability required for systematic studies. Our next step is to add feedback control using force estimated by the model from the senor readings, so that the robot can adjust body bending to better conform to and push against complex 3-D terrain [54].

## ACKNOWLEDGMENT

We thank Xiangyu Peng, Nikhil Murty and Kaiwen Wang for early sensor and robot prototyping; Yaqing Wang for calibration setup development; and Yaqing Wang, Ratan Othayoth, and Henry Astley for discussion. D.R. developed robot, conducted experiments, analyzed data, and developed model; Q.F. developed robot control and assisted robot design and experimental setup; D.R. and C.L wrote the paper.